\documentclass[11pt]{article}

\usepackage[final]{acl}

\setcitestyle{numbers,square}

\usepackage{latexsym}
\usepackage{hyperref}

\usepackage{markdown}

\usepackage{fontspec}
\defaultfontfeatures{Ligatures=TeX}
\newfontfamily\devanagarifont{NotoSerifDevanagari-VariableFont.ttf}[Script=Devanagari]
\newfontfamily\kannadafont{NotoSerifKannada-VariableFont.ttf}[Script=Kannada]

\usepackage{ucharclasses}
\setTransitionsFor{Devanagari}{\devanagarifont}{\normalfont}
\setTransitionsFor{Kannada}{\kannadafont}{\normalfont}
\setTransitionTo{Latin}{\normalfont}

\newcommand{\dn}[1]{{\devanagarifont #1}}
\newcommand{\kn}[1]{{\kannadafont #1}}

\usepackage{microtype}

\usepackage{inconsolata}

\usepackage{listings}
\usepackage{fvextra}
\DefineVerbatimEnvironment{WrappedVerbatim}{Verbatim}{%
  breaklines,breakanywhere,
  breaksymbolleft={},breaksymbolright={}, % remove the "↪" continuation marker
  xleftmargin=0pt,
  fontsize=\footnotesize%
}
\DefineVerbatimEnvironment{WrappedVerbatimKN}{Verbatim}{%
  breaklines,breakanywhere,
  breaksymbolleft={},breaksymbolright={},
  xleftmargin=0pt,
  fontsize=\footnotesize,
  formatcom=\ttfamily%
}

\usepackage[most]{tcolorbox}
\tcbuselibrary{skins,listings,breakable}

\newtcblisting{systemprompt}[1][]{
  skin=bicolor,
  colback=gray!5,
  colframe=gray!40,
  colbacktitle=gray!20,
  coltitle=black,
  fonttitle=\bfseries\sffamily,
  title={System Prompt Context},
  listing only,
  listing options={
    basicstyle=\ttfamily\scriptsize,
    breaklines=true,
    columns=fullflexible,
    keepspaces=true,
    upquote=true,
    tabsize=2,
    literate={*}{{\char42}}1 {-}{{\char45}}1,
  },
  sharp corners,
  boxrule=0.5pt,
  left=5pt, right=5pt, top=5pt, bottom=5pt,
  breakable,
  enhanced,
  #1
}

\usepackage{graphicx}
\usepackage{booktabs}
\usepackage{array}
\usepackage{tabularx}
\usepackage{longtable}
\usepackage{multirow}
\usepackage{pgfplots}
\pgfplotsset{compat=1.18}
\usepackage{placeins} % provides \FloatBarrier to stop floats crossing sections
\usepackage{float} % provides [H] to place a float exactly here
\usepackage{needspace} % keep section headings with following text

\title{Konkani LLM: Multi-Script Instruction Tuning and Evaluation for a Low-Resource Indian Language}

\author{
    Reuben Chagas Fernandes\textsuperscript{1},
    Gaurang S. Patkar\textsuperscript{1}
    \\
    \textsuperscript{1}Don Bosco College Of Engineering, Fatorda
}

\begin{document}
\maketitle
\begin{abstract}
Large Language Models (LLMs) consistently under perform in low-resource linguistic contexts such as Konkani. This performance deficit stems from acute training data scarcity compounded by high script diversity across Devanagari, Romi and Kannada orthographies. To address this gap, we introduce Konkani-Instruct-100k, a comprehensive synthetic instruction-tuning dataset generated through Gemini 3.

We establish rigorous baseline benchmarks by evaluating leading open-weights architectures including Llama 3.1, Qwen2.5 and Gemma 3 alongside proprietary closed-source models. Our primary contribution involves the development of Konkani LLM, a series of fine-tuned models optimized for regional nuances. Furthermore, we are developing the Multi-Script Konkani Benchmark to facilitate cross-script linguistic evaluation. In machine translation, Konkani LLM delivers consistent gains over the corresponding base models and is competitive with and in several settings surpasses proprietary baselines (Table~\ref{tab:auto-metrics-translation}).
\end{abstract}

\section{Introduction}
\subsection{Background}
Konkani is an Indo-Aryan language distinct to the Goa and Konkan regions of India. It presents substantial challenges for Natural Language Processing (NLP) due to its multiscript nature. The language uses Devanagari, Romi (Latin) and Kannada scripts officially. Konkani retains low-resource status despite its recognition as a scheduled language of India. High-quality parallel corpora remain scarce. Existing datasets suffer from fragmentation across these diverging scripts which impedes the training of unified language models.

\subsection{The Gap: Language Contamination}
Large language models (LLMs) such as Deepseek V3.2, Claude sonnet 4.6 and GPT-5.2 achieve strong multilingual performance but Konkani remains challenging. While some base models may have encountered Konkani during pre-training, they typically lack sufficient instruction-tuning data to reliably treat Konkani as a distinct target language.

\subsection{Proposed Solution}
We adopt a pipeline centered on Synthetic Data Generation  and parameter-efficient fine-tuning to mitigate these issues. We used Gemini 3 to synthesize Konkani-Instruct-100k in the absence of native instruction-tuning data. The dataset balances Devanagari, Romi and Kannada scripts. We then fine-tune open-weight architectures, including Llama 3.1 8B, Gemma 3 4B, Gemma 3 12B, Gemma 3 27B, Qwen2.5 14B and Qwen2.5 1.5B on this synthetic corpus. Our goal is to show that compact models trained on linguistically controlled data can outperform general-purpose counterparts in script fidelity and grammatical accuracy for Konkani.

\subsection{Contributions}
This study delivers the following contributions:

Konkani-Instruct-100k: The first large-scale multiscript instruction-tuning dataset generated via distillation.

Konkani LLM: A suite of fine-tuned models optimized for Konkani.

Konkani-Bench: A 200-pair benchmark spanning Romi, Devanagari and Kannada scripts alongside English.

\section{Related Work}
Research on large language models for Indian languages has accelerated, yet coverage remains skewed towards high-resource languages such as Hindi and Tamil, with comparatively limited attention to Konkani. OpenHathi extends Llama 2 7B with an expanded tokenizer and bilingual training to support Hindi, English and Hinglish and reports performance comparable to GPT-3.5 on Hindi tasks in both Devanagari and Roman scripts~\citep{openhathi}. Sarvam-1 targets ten Indic languages and reports strong results on benchmarks such as MMLU, ARC-Challenge, IndicGenBench and Flores translation and includes support for Konkani; however, it does not explicitly address Konkani in Roman (Romi) and Kannada scripts~\citep{sarvam1}. Tamil-LLaMA augments LLaMA 2 with additional Tamil tokens and instruction tuning on translated instruction datasets to improve monolingual Tamil understanding and generation, but the approach is not extended to minority Goan languages~\citep{tamilllama}. Existing Indic LLM efforts therefore provide strong baselines for Hindi and Tamil but leave a gap for low-resource languages like Konkani, especially in code-mixed, multi-script settings.

Synthetic instruction data has emerged as a standard strategy to compensate for limited supervised corpora and to boost downstream performance. Self-Instruct style pipelines use strong teacher models to generate diverse instructions and responses, followed by automatic filtering and curriculum design for student model fine-tuning~\citep{selfinstruct}. CoT-Self-Instruct extends this paradigm by enforcing chain-of-thought planning before synthetic example generation and reports higher accuracy on reasoning tasks compared to vanilla Self-Instruct~\citep{cotselfinstruct}. These methods indicate that synthetic data generation is a viable approach for building instruction-tuned models in settings where high-quality human-annotated data for Konkani is sparse or absent.

Work on script handling for Indian languages covers both transliteration and parallel data construction across scripts. Tamil-LLaMA relies on a standardized Tamil script and improved subword segmentation rather than explicit multi-script modeling, which limits transfer to languages like Konkani that routinely appear in Devanagari, Roman and Kannada scripts~\citep{tamilllama}. For Konkani specifically, existing NLP research focuses on foundational tasks such as POS tagging and sentiment analysis, often by leveraging transfer from resource-rich Marathi through parallel corpora and zero-shot models, not through dedicated multi-script LLMs. Recent efforts on Indic LLMs and resources highlight the scarcity of digital resources for Konkani and motivate dedicated multi-script evaluation and modeling~\citep{ai4bharat}.

% TODO: Add BibTeX entries for \citep{openhathi,sarvam1,tamilllama,selfinstruct,cotselfinstruct,ai4bharat} in latex/custom.bib.

\section{Konkani-Instruct-100k}
\subsection{Dataset Creation}

\subsubsection{Preliminary Experiments}
Initial resource scarcity necessitated reliance on sparse open-source repositories. We utilized a 15,000-sample subset from Nagvekar~\citep{nagvekar_asmitai_noisified}, consisting of Wikipedia-based edit/fix tasks rather than instruction-following pairs. We augmented these data with a Devanagari Alpaca dataset and employed Aksharamukha for transliteration into Romi and Kannada scripts to address script scarcity.

Pilot fine-tuning revealed limited generalization: models produced poorly formed sentences and struggled with open-ended queries. We attribute this to two factors. First, the closed-style nature of the training data did not promote conversational fluency. Second, automated transliteration introduced phonological errors that conflicted with dialectal nuances.

\subsubsection{Synthetic Methodology}
A synthetic generation pipeline addressed these structural weaknesses. Based on human evaluation, we selected Gemini 2.5 Pro and Gemini 3 Flash as the most reliable teacher models for Konkani. We also tested Grok 4.1, which performed competitively but tended to produce less appropriate lexical choices and occasionally generated NSFW content, making it unsuitable for data generation. We implemented a "Tutor-Style" pedagogical framework rather than standard QA pairs. This approach teaches core linguistic mechanics via 18 foundational topics (Table~\ref{tab:foundational-topics}) to ensure basic competence in everyday Konkani.

\subsubsection{Data Structure and Morphology}
Morphological analysis tables constitute a distinguishing feature of this dataset. All 100\% of samples provide the answer alongside a structural breakdown including part-of-speech tags and base-modifier morphology. Over 3,000 samples explicitly detail grammatical rules such as gender suffix modifications. Alternative phrasings for identical intents mitigate overfitting.

\subsubsection{Prompting and Domain Diversity}
Beyond these foundational lessons, we expand coverage by varying prompt domains (e.g., politics, science and technology) to increase contextual diversity. We employ a human-in-the-loop process in which the teacher model populates reusable instruction skeletons with synthetic entities, yielding prompts that better reflect natural variation.

\begin{table}[t]
\centering
\small
\begin{tabular}{r l}
\toprule
\textbf{ID} & \textbf{Foundational topic} \\
\midrule
1  & Food \\
2  & Family relationships \\
3  & Numbers \\
4  & Time \\
5  & Body parts \\
6  & Colors \\
7  & Nature \\
8  & Common words \\
9  & Animals \\
10 & Adjectives \\
11 & Pronouns \\
12 & Tenses \\
13 & Question formation \\
14 & Gender \\
15 & Locations \\
16 & Names \\
17 & Greetings \\
18 & Oblique cases and active voice \\
\bottomrule
\end{tabular}
\caption{Foundational topics used to generate tutor-style lessons for basic Konkani competence.}
\label{tab:foundational-topics}
\end{table}

\begin{table}[t]
\centering
\small
\begin{tabular}{lrr}
\toprule
\textbf{Script} & \textbf{Train} & \textbf{Validation} \\
\midrule
Devanagari & 26,006 & 4,820 \\
Romi (Latin) & 38,382 & 4,008 \\
Kannada & 26,845 & 3,917 \\
English & 1,000 & 1,000 \\
\midrule
\textbf{Total} & 92,233 & 13,745 \\
\bottomrule
\end{tabular}
\caption{Distribution of Konkani-Instruct-100k across scripts and splits.}
\label{tab:dataset-distribution}
\end{table}

\subsubsection{Dataset Statistics}
The final Konkani-Instruct-100k dataset comprises 105,982 samples, balanced across scripts to reduce orthographic bias. We additionally include a small set of English-to-English examples derived from the Sentence-Transformers Natural Questions dataset \citep{st_naturalquestions} as a regularizer to help preserve general-purpose reasoning.

\section{Konkani Multi-Script Benchmark}
We are developing a human-annotated benchmark for translation and transliteration, intended to stress-test (i) script fidelity and (ii) cross-script robustness for Konkani.

\subsection{Task suite}
The benchmark is built from 200 human-annotated items. For each item, we create parallel variants in Devanagari, Romi and Kannada and include an English translation to enable bilingual evaluation. We structure evaluation around two task families:
\begin{itemize}
    \item \textbf{Translation:} Konkani-to-English.
    \item \textbf{Transliteration:} pairwise transliteration between Romi, Devanagari and Kannada (Romi--Devanagari, Romi--Kannada and Devanagari--Kannada; both directions).
\end{itemize}

\subsection{Data splits and scripts}
Each item is provided in Devanagari, Romi and Kannada along with an English translation. We recommend reporting results per-script and macro-averaged across scripts.

\subsection{Evaluation protocol}
We use automatic metrics for translation quality and transliteration accuracy and supplement them with human review for script fidelity and lexical appropriateness.

% TODO: Describe benchmark size, sources, licensing, and provide examples in an appendix.

\section{Methodology}
We fine-tune open-weight base models on Konkani-Instruct-100k using parameter-efficient adaptation.

\subsection{Data generation and filtering}
We generate instruction--response pairs with a teacher model and apply automatic filters to reduce malformed or repetitive samples. We then balance scripts during sampling to avoid overfitting to a single orthography.

\subsection{Training approach}
We use Low-Rank Adaptation (LoRA) for parameter-efficient fine-tuning. We choose LoRA because it allows us to release lightweight adapter weights and support low-cost deployment by uploading adapters to hosted inference providers.

\paragraph{LoRA configuration.}
We set rank $r=64$ and scaling $\alpha=128$, apply LoRA to attention projections (\texttt{q\_proj}, \texttt{k\_proj}, \texttt{v\_proj}, \texttt{o\_proj}) and MLP projections (\texttt{gate\_proj}, \texttt{up\_proj}, \texttt{down\_proj}), use dropout 0 and set bias adaptation to \texttt{none}. We enable gradient checkpointing for long-context training.

\paragraph{Optimization and checkpointing.}
Unless otherwise stated, we fine-tune for 2 epochs with maximum sequence length 4096 and select checkpoints by validation loss. We train with per-device batch size 1 and gradient accumulation 32 (effective batch size 32), 8-bit AdamW, a cosine learning-rate schedule with 10\% warmup and bfloat16 precision. We evaluate and checkpoint every 500 steps, keep the best checkpoint by validation loss and cap the number of saved checkpoints at 3.

% TODO: Report any quantization (e.g., QLoRA/4-bit), max steps vs epoch-based stopping details, data packing strategy, and the exact train/validation preprocessing pipeline.

\section{Experimental Setup \& Metrics}

% TODO (future work): add a table of proprietary (closed-source) models once evaluated.

% TODO (future work): add a table of open-weight models once evaluation is finalized.

% TODO: Add parameter counts, context length, and decoding settings (temperature/top-$p$/max tokens) for each model.
\subsection{Training configuration}
\begin{table}[H]
\centering
\small
\setlength{\tabcolsep}{4pt}
\renewcommand{\arraystretch}{1.1}
\sloppy
\begin{tabularx}{\linewidth}{lX}
\toprule
\textbf{Setting} & \textbf{Value (default)} \\
\midrule
Sequence length & 4096 \\
Epochs & 2 \\
Per-device batch size & Train 1 (some runs: 4), Eval 1 \\
Gradient accumulation & 32 (some runs: 8; effective batch size 32) \\
Optimizer & AdamW 8-bit (\texttt{adamw\_8bit}) \\
Learning rate & $2\times 10^{-4}$ \\
Scheduler / warmup & Cosine; warmup ratio 0.1 \\
Weight decay / grad clip & 0.01 / 1.0 \\
Precision & bfloat16 \\
Eval / save cadence & Every 500 steps; keep best by \texttt{eval\_loss}; save limit 3 \\
LoRA rank / $\alpha$ & $r=64$, $\alpha=128$ \\
LoRA target modules & \texttt{q\_proj, k\_proj, v\_proj, o\_proj, gate\_proj, up\_proj, down\_proj} \\
LoRA dropout / bias & 0 / \texttt{none} \\
Gradient checkpointing & Enabled via Unsloth (long-context) \\
Seed & 3407 \\
\bottomrule
\end{tabularx}
\caption{Default fine-tuning configuration used across models unless otherwise stated.}
\label{tab:ft-config}
\end{table}

We use the default fine-tuning configuration in Table~\ref{tab:ft-config} unless otherwise stated.

% TODO: Add exact number of update steps, optimizer betas/epsilon, any quantization (e.g., QLoRA/4-bit), packing strategy, and total wall-clock time.

\FloatBarrier % ensure Table~\ref{tab:ft-config} appears before Metrics
\subsection{Metrics}
We evaluate models using (i) reference-based automatic metrics: BLEU~\cite{bleu}, chrF++~\cite{chrf} and COMET for translation (Table~\ref{tab:auto-metrics-translation}) and BLEU and chrF++ for transliteration (Table~\ref{tab:auto-metrics-transliteration}). We also use (ii) an LLM-as-a-judge with a fixed rubric that scores helpfulness, script fidelity and absence of Marathi/Hindi contamination; unless otherwise stated, we use Gemini 3 Flash as the judge model (Tables~\ref{tab:judge-simple-creative}--\ref{tab:judge-overall}). We note that IndicParam \citep{indicparam} includes Konkani multiple-choice questions, but we do not use it because it targets only Devanagari and is less informative for measuring cross-script robustness and translation quality.

% Reference-based automatic metrics (translation vs. transliteration)

\begin{table*}[!t]
\centering
\small 
\setlength{\tabcolsep}{8pt} % Increased slightly for better spacing with fewer columns
\renewcommand{\arraystretch}{1.1}
\begin{tabular}{l rrr}
\toprule
\textbf{Model} & \textbf{BLEU} & \textbf{chrF++} & \textbf{COMET} \\
\midrule
\multicolumn{4}{l}{\textit{Proprietary \& Large API Models}} \\
gemini-flash-latest              & 86.76 & 92.13 & 91.14 \\
gemini-3-pro-preview             & 71.50 & 83.73 & 88.56 \\
gpt-5.2                          & 57.37 & 74.04 & 85.93 \\
claude-sonnet-4-6                & 57.16 & 74.65 & 84.43 \\
gpt-5-mini                       & 34.87 & 54.76 & 75.53 \\
grok-4-fast-non-reasoning        & 29.38 & 51.22 & 71.72 \\
kimi-k2-instruct-0905            & 46.40 & 64.05 & 80.41 \\
\midrule
\multicolumn{4}{l}{\textit{Open-Weights (General)}} \\
DeepSeek-V3.2                    & 46.05 & 63.11 & 80.91 \\
gemma-3-27b-it                   & 32.64 & 52.59 & 73.87 \\
Llama-3.3-70B-Instruct           & 24.74 & 45.11 & 69.97 \\
gemma-3-12b-it                   & 18.18 & 39.21 & 66.30 \\
command-a-03-2025                & 1.92  & 18.66 & 59.18 \\
c4ai-aya-expanse-32b             & 6.27  & 28.24 & 58.13 \\
Meta-Llama-3.1-8B-Instruct       & 6.85  & 28.20 & 57.10 \\
Qwen2.5-14B-Instruct             & 5.59  & 25.77 & 56.90 \\
Qwen2.5-1.5B-Instruct            & 2.11  & 19.56 & 49.88 \\
\midrule
\multicolumn{4}{l}{\textit{Konkani Finetuned Models}} \\
konkani-Qwen2.5-14B-Instruct     & 50.26 & 68.98 & 83.58 \\
konkani-gemma-3-27b-it           & 48.76 & 67.67 & 82.91 \\
konkani-gemma-3-12b-it           & 41.96 & 62.48 & 80.48 \\
konkani-llama3.1-8b   & 37.60 & 62.23 & 79.18 \\
konkani-gemma-3-4b-it            & 34.10 & 58.02 & 79.08 \\
konkani-Qwen2.5-1.5B-Instruct    & 2.68  & 14.61 & 36.94 \\
\bottomrule
\end{tabular}
\caption{Automatic metrics for translation (200 samples). The ``Task'' column has been omitted as all entries represent translation tasks. Models are grouped by category and sorted by COMET score descending within groups.}
\label{tab:auto-metrics-translation}
\end{table*}

\begin{table*}[!t]
\centering
\scriptsize
\setlength{\tabcolsep}{3pt}
\renewcommand{\arraystretch}{1.1}
\resizebox{\textwidth}{!}{%
\begin{tabular}{l rrrrrr rrrrrr}
\toprule
\textbf{Model} & \multicolumn{6}{c}{\textbf{BLEU}} & \multicolumn{6}{c}{\textbf{chrF++}} \\
\cmidrule(lr){2-7} \cmidrule(lr){8-13}
& D$\to$K & D$\to$R & K$\to$D & K$\to$R & R$\to$D & R$\to$K & D$\to$K & D$\to$R & K$\to$D & K$\to$R & R$\to$D & R$\to$K \\
\midrule
gemini-flash-latest            & 29.83 & 43.83 & 35.07 & 47.66 & 78.13 & 67.01 & 66.69 & 77.34 & 75.90 & 79.59 & 91.86 & 87.70 \\
claude-sonnet-4-6              & 21.11 & 10.48 & 23.75 & 10.08 & 44.77 & 30.75 & 58.32 & 43.42 & 57.82 & 39.51 & 72.86 & 67.60 \\
gpt-5.2                        & 7.79  & 13.35 & 33.46 & 12.58 & 31.89 & 15.09 & 44.76 & 46.70 & 67.02 & 47.76 & 66.52 & 56.03 \\
gpt-5-mini                     & 1.95  & 6.57  & 20.41 & 9.13  & 11.74 & 5.90  & 33.89 & 43.54 & 57.81 & 48.53 & 47.84 & 41.16 \\
grok-4-fast-non-reasoning      & 8.28  & 7.87  & 13.30 & 10.66 & 6.81  & 4.16  & 44.47 & 41.92 & 52.12 & 47.53 & 42.50 & 37.78 \\
kimi-k2-instruct-0905          & 9.08  & 8.51  & 26.74 & 9.96  & 16.18 & 9.92  & 44.46 & 42.41 & 63.40 & 46.56 & 53.25 & 48.24 \\
\midrule
DeepSeek-V3.2                  & 22.62 & 12.24 & 31.85 & 11.63 & 32.91 & 10.97 & 57.98 & 49.47 & 65.64 & 49.89 & 65.02 & 47.77 \\
gemma-3-27b-it                 & 13.99 & 11.83 & 25.57 & 16.56 & 14.91 & 5.96  & 51.20 & 49.53 & 62.14 & 56.03 & 49.20 & 41.66 \\
gemma-3-12b-it                 & 7.10  & 4.84  & 11.71 & 10.40 & 5.05  & 2.94  & 37.20 & 38.78 & 41.54 & 47.88 & 31.62 & 30.48 \\
Llama-3.3-70B-Instruct         & 13.82 & 7.08  & 28.91 & 10.00 & 24.57 & 12.22 & 52.29 & 42.60 & 65.32 & 48.06 & 60.10 & 52.19 \\
Meta-Llama-3.1-8B-Instruct     & 4.92  & 4.72  & 25.31 & 5.44  & 10.72 & 2.30  & 37.12 & 37.35 & 61.59 & 37.35 & 44.13 & 28.58 \\
c4ai-aya-expanse-32b           & 1.30  & 2.92  & 11.36 & 4.27  & 2.50  & 1.13  & 29.70 & 34.85 & 48.95 & 37.30 & 32.72 & 29.30 \\
command-a-03-2025              & 5.09  & 0.19  & 5.57  & 2.79  & 0.32  & 2.74  & 39.42 & 10.34 & 34.90 & 28.41 & 14.91 & 24.81 \\
Qwen2.5-14B-Instruct           & 0.66  & 1.30  & 10.26 & 2.11  & 1.90  & 0.23  & 25.66 & 21.40 & 48.49 & 27.87 & 27.09 & 17.28 \\
Qwen2.5-1.5B-Instruct          & 0.08  & 0.30  & 1.21  & 0.27  & 0.27  & 0.07  & 7.88  & 8.46  & 10.69 & 9.55  & 13.79 & 5.88  \\
\midrule
konkani-Qwen2.5-14B-Instruct   & 37.51 & 50.52 & 56.03 & 55.26 & 55.35 & 43.95 & 72.25 & 80.04 & 79.66 & 81.41 & 78.35 & 74.99 \\
konkani-gemma-3-27b-it         & 38.71 & 45.81 & 15.91 & 50.11 & 55.02 & 38.38 & 72.37 & 76.41 & 49.03 & 78.92 & 78.29 & 71.66 \\
konkani-gemma-3-12b-it         & 19.09 & 48.56 & 3.51  & 45.05 & 27.92 & 31.85 & 62.80 & 78.07 & 23.55 & 78.49 & 61.43 & 67.72 \\
konkani-gemma-3-4b-it          & 22.05 & 42.19 & 1.43  & 36.49 & 4.52  & 7.82  & 61.52 & 70.41 & 12.79 & 64.18 & 23.15 & 42.72 \\
konkani-llama3.1-8b & 26.88 & 30.36 & 19.30 & 23.43 & 21.93 & 22.12 & 69.69 & 74.13 & 49.02 & 70.13 & 63.81 & 66.86 \\
konkani-Qwen2.5-1.5B-Instruct  & 8.95  & 0.97* & 0.23  & 1.52  & 14.58 & 11.64 & 39.91 & 13.10*& 2.06  & 12.40 & 35.33 & 36.97 \\
\bottomrule
\end{tabular}%
}
\caption{Automatic metrics for transliteration. Directions are abbreviated as follows: \textbf{D} (Devanagari), \textbf{K} (Kannada), \textbf{R} (Romi). Note that COMET is excluded as it is not applicable to transliteration. All pairs evaluated on 200 samples except where marked with * (160 samples).}
\label{tab:auto-metrics-transliteration}
\end{table*}

% LLM-as-a-judge tables (placed after Tables~\ref{tab:auto-metrics-translation} and~\ref{tab:auto-metrics-transliteration})
\begin{table*}[!t]
\centering
\small
\setlength{\tabcolsep}{4pt}
\renewcommand{\arraystretch}{1.1}
\begin{tabular}{lccc ccc}
\toprule
Base model & \multicolumn{3}{c}{Simple chat} & \multicolumn{3}{c}{Creative writing} \\
\cmidrule(lr){2-4}\cmidrule(lr){5-7}
 & Romi & Devanagari & Kannada & Romi & Devanagari & Kannada \\
\midrule
Qwen2.5 1.5B & 2.3 & 3.1 & 3.9 & 2.1 & 1.3 & 1.9 \\
Qwen2.5 14B & 4.0 & 4.4 & 4.4 & 3.7 & 2.8 & 3.6 \\
Gemma 4B & 3.4 & 4.2 & 3.1 & 1.9 & 1.9 & 2.0 \\
Gemma 12B & 3.9 & 3.2 & 4.5 & 2.5 & 2.2 & 1.9 \\
Gemma 27B & 4.6 & 4.5 & 4.3 & 2.8 & 3.2 & 3.0 \\
Llama 8B & 4.6 & 4.6 & 4.3 & 4.2 & 3.7 & 2.6 \\
\bottomrule
\end{tabular}
\caption{LLM-as-a-judge scores (1--5) for Simple chat and Creative writing by script (judge: Gemini 3 Flash).}
\label{tab:judge-simple-creative}
\end{table*}

\begin{table*}[!t]
\centering
\small
\setlength{\tabcolsep}{4pt}
\renewcommand{\arraystretch}{1.1}
\begin{tabular}{lccc ccc}
\toprule
Base model & \multicolumn{3}{c}{Logical reasoning} & \multicolumn{3}{c}{Translation} \\
\cmidrule(lr){2-4}\cmidrule(lr){5-7}
 & Romi & Devanagari & Kannada & Romi & Devanagari & Kannada \\
\midrule
Qwen2.5 1.5B & 2.0 & 1.6 & 2.1 & 2.8 & 2.2 & 2.5 \\
Qwen2.5 14B & 3.9 & 3.5 & 3.4 & 4.5 & 4.3 & 4.8 \\
Gemma 4B & 2.4 & 2.8 & 2.8 & 3.1 & 3.5 & 3.1 \\
Gemma 12B & 3.0 & 3.3 & 2.4 & 4.0 & 3.7 & 4.0 \\
Gemma 27B & 3.3 & 4.4 & 3.2 & 3.5 & 3.5 & 1.8 \\
Llama 8B & 4.5 & 4.5 & 4.2 & 4.3 & 4.8 & 3.2 \\
\bottomrule
\end{tabular}
\caption{LLM-as-a-judge scores (1--5) for Logical reasoning and Translation by script (judge: Gemini 3 Flash).}
\label{tab:judge-logic-translation}
\end{table*}

\begin{table*}[!t]
\centering
\small
\setlength{\tabcolsep}{6pt}
\renewcommand{\arraystretch}{1.1}
\makebox[\textwidth][c]{%
\begin{tabular}{lccc}
\toprule
Base model & Romi & Devanagari & Kannada \\
\midrule
Qwen2.5 1.5B & 2.30 & 2.05 & 2.60 \\
Qwen2.5 14B & 4.03 & 3.75 & 4.05 \\
Gemma 4B & 2.70 & 3.10 & 2.75 \\
Gemma 12B & 3.35 & 3.10 & 3.20 \\
Gemma 27B & 3.55 & 3.90 & 3.08 \\
Llama 8B & 4.40 & 4.40 & 3.58 \\
\bottomrule
\end{tabular}%
}
\caption{Overall LLM-as-a-judge scores (1--5) by script (judge: Gemini 3 Flash).}
\label{tab:judge-overall}
\end{table*}

% TODO: State judge prompt, decoding settings, and inter-run stability checks.
% TODO (future): Add reference-based automatic metrics once the evaluation pipeline is finalized.
% TODO: State judge prompt, decoding settings, and inter-run stability checks.

% TODO (future): Add reference-based automatic metrics once the evaluation pipeline is finalized.
% TODO: State judge prompt, decoding settings, and inter-run stability checks.

\section{Evaluation}
\subsection{Benchmark construction}
We evaluate models on Konkani-Bench \citep{konkani_bench}, a Konkani multi-script benchmark constructed as follows: we first write human-authored Romi Konkani sentences, then (i) transliterate them into Devanagari and Kannada and (ii) translate them into English to create references.

\subsection{Automatic metrics}
For translation, we report BLEU~\cite{bleu}, chrF++~\cite{chrf}, and COMET (Table~\ref{tab:auto-metrics-translation}). For transliteration, we report BLEU and chrF++ across script directions (Table~\ref{tab:auto-metrics-transliteration}).

\subsection{LLM-as-a-judge evaluation}
We additionally evaluate outputs using an LLM-as-a-judge with a fixed rubric that scores helpfulness, script fidelity  and absence of Marathi/Hindi contamination; unless otherwise stated, we use Gemini 3 Flash as the judge model (Tables~\ref{tab:judge-simple-creative}--\ref{tab:judge-overall}).

% TODO: Add qualitative examples and error categories in Appendix D.

\FloatBarrier
\section{Limitations}
Our dataset is largely synthetic and may encode teacher-model biases, stylistic artifacts, or factual inaccuracies. While script balancing reduces orthographic bias, it cannot fully address dialect diversity across Konkani communities. Additionally, our deployment constraints required using LoRA~\cite{lora} with rank $r=64$ to support serverless LoRA inference, which may underperform higher-rank adapters in some settings. Finally, due to time and budget constraints, we did not experiment with continued pre-training or full fine-tuning; we expect that these heavier training regimes would further improve performance beyond what we report here.

\needspace{6\baselineskip}
\section{Conclusion}
This work takes a step toward stronger multi-script support for Konkani by introducing Konkani-Instruct-100k and fine-tuned Konkani LLM variants. On reference-based translation evaluation (Table~\ref{tab:auto-metrics-translation}), our best-performing fine-tuned model is \texttt{konkani-Qwen2.5-14B-Instruct}, achieving the highest COMET score among our models. For transliteration (Table~\ref{tab:auto-metrics-transliteration}), our fine-tuned models substantially improve over their corresponding bases and show strong cross-script robustness across directions. Using LLM-as-a-judge evaluation (Tables~\ref{tab:judge-simple-creative}--\ref{tab:judge-overall}), Llama~8B performs best overall for Romi and Devanagari, while Qwen2.5~14B performs best for Kannada. We release models on Hugging Face at \href{https://huggingface.co/konkani}{huggingface.co/konkani}, and provide a public demo at \href{https://konkani.app/}{konkani.app} for easy try-out.

\needspace{6\baselineskip}
\section*{Acknowledgments}
\begin{large}
We thank \href{https://www.cloudrift.ai/}{Cloud Riff}, the Hugging Face community, and everyone who supported this project. We also thank the Cohere Labs community for helpful discussions, with special thanks to Alejandro Salamanca. We are grateful to Prof. Amey Kerker (co-guide) and Prof. Mithil Parab (reviewer) for their guidance and feedback. Finally, we thank \href{https://modal.com/}{Modal} for providing access to H200 GPUs and compute that enabled this work.
\end{large}

\clearpage
\onecolumn
\appendix

\section{Appendix A: Data Generation Prompts}
\label{sec:appendix_prompts}

This appendix provides the specific prompt templates used for evaluating the models on different tasks.

\subsection{Roman Konkani (Romi) translation prompt (Gemini 3 Flash)}
\noindent\textbf{Note:} Gemini~3~Flash is a \textit{thinking} model.

% Usage: \begin{systemprompt}[Title=...] content \end{systemprompt}
\begin{systemprompt}[title=System Prompt: Romi Konkani Tutor]
**Role:** You are an expert Linguistic Tutor specializing in **Roman Konkani (Roman Catholic Konkani)**. Your objective is to translate English sentences into grammatically perfect Konkani by strictly deconstructing the morphology (Gender, Politeness, Case) before generating the final output.

**Core Directive:**
You must never output a translation immediately. You must first prove your reasoning using a **Grammar Decomposition Table**. This prevents hallucinations regarding gender agreement and oblique cases.

### 1. Analysis Protocols

Before generating text, you must internally resolve the following:

* **Gender:** Every noun must be tagged as **[Masculine]** (Masculine), **[Feminine]** (Feminine), or **[Neuter]** (Neuter).
* **Politeness:** Determine if the listener is **Informal (Tu)** or **Formal/Plural (Tumi)**.
* **Oblique Cases:** If a noun is followed by a preposition (in, on, at, to), you must identify the **Oblique Base** (e.g., *Ghar* -> *Ghora*) before adding the suffix.

### 2. Output Format (Strict)

Every response must follow this exact structure:

**A. The Grammar Decomposition Table**
Create a Markdown table with the following specific columns:

* `Role`: (Subject, Object, Adjective, Verb, etc.)
* `English`: The source word.
* `Konkani Root`: The dictionary form of the word.
* `Tags`: Gender [Masculine/Feminine/Neuter], Number [Singular/Plural], Politeness [Informal/Formal].
* `Morphology Logic`: The rule applied (e.g., "Masculine noun ending in 'o' becomes 'a' in oblique case").
* `Final Word`: The conjugated/modified word.

**B. The Final Translation**
The assembled sentence, bolded.

### 3. Rules of Grammar (The "Truth Source")

* **Adjective Agreement:** Adjectives must match the Noun's gender.
  * *Mhojo* (Masculine), *Mhoji* (Feminine), *Mhojem* (Neuter).

* **Pronoun-Verb Consistency:**
  * *Tu* (You-Informal) -- Verb usually ends in *-ta/-la*.
  * *Tumi* (You-Formal) -- Verb usually ends in *-tat/-le*.

* **Oblique Case Trigger:** Prepositions are suffixes in Konkani.
  * *Table* (Mez [Neuter]) + *On* (-r) -- Do not say *Mez-r*. Convert to oblique (*Mez* -> *Meza*) -> **Mezar**.

---

### Example of Required Output

**User Input:** "My younger sister is playing in the garden."

**Model Response:**

| Role | English | Konkani Root | Tags | Morphology Logic | Final Word |
| --- | --- | --- | --- | --- | --- |
| **Possessive** | My | Mhojo | [Feminine] Singular | Matches "Bhoinn" (Sister) | **Mhoji** |
| **Adjective** | Younger | Lhan | [Feminine] Singular | Adjective agrees with Noun | **Lhan** |
| **Subject** | Sister | Bhoinn | [Feminine] Singular | Root remains subject | **Bhoinn** |
| **Object** | Garden | Mollo | [Masculine] Singular | **Oblique Rule:** Masc '-o' becomes '-ea' | **Mollea** |
| **Postposition** | In | -nt | - | Added to oblique base | **Molleant** |
| **Verb** | Is playing | Khell- | Present continuous. | Matches "Ti" (She) form | **Khellta** |

**Final Sentence:**

> **[final sentence in Roman Catholic Konkani (Romi) script]**

Grammatical Insights
* Explain *why* a specific pronoun was selected.
* Explain how the Gender/Number of the object forced the Adjective or Verb to change.
* Explain the Oblique Case rule and why it was applied to the object.
* Identify the trigger (Preposition or Postposition) that caused the noun to bend.
\end{systemprompt}

\subsection{Additional prompt templates}

\subsubsection{Devanagari Konkani (Standard script) translation prompt}

\begin{systemprompt}[title=System Prompt: Devanagari Konkani Tutor]
**Role:** You are an expert Linguistic Tutor specializing in **Devanagari Konkani (Standard script)**. Your objective is to translate English sentences into grammatically perfect Konkani by strictly deconstructing the morphology (Gender, Politeness, Case) before generating the final output.

**Core Directive:**
You must never output a translation immediately. You must first prove your reasoning using a **Grammar Decomposition Table**. This prevents hallucinations regarding gender agreement and oblique cases.

### 1. Analysis Protocols

Before generating text, you must internally resolve the following:

* **Gender:** Every noun must be tagged as **[Masculine]** (Masculine), **[Feminine]** (Feminine), or **[Neuter]** (Neuter).
* **Politeness:** Determine if the listener is **Informal (Tu)** or **Formal/Plural (Tumi)**.
* **Oblique Cases:** If a noun is followed by a preposition (in, on, at, to), you must identify the **Oblique Base** (e.g., *Ghar* -> *Ghora*) before adding the suffix.

### 2. Output Format (Strict)

Every response must follow this exact structure:

**A. The Grammar Decomposition Table**
Create a Markdown table with the following specific columns:

* `Role`: (Subject, Object, Adjective, Verb, etc.)
* `English`: The source word.
* `Konkani Root`: The dictionary form of the word.
* `Tags`: Gender [Masculine/Feminine/Neuter], Number [Singular/Plural], Politeness [Informal/Formal].
* `Morphology Logic`: The rule applied (e.g., "Masculine noun ending in 'o' becomes 'a' in oblique case").
* `Final Word`: The conjugated/modified word.

**B. The Final Translation**
The assembled sentence, bolded.

### 3. Rules of Grammar (The "Truth Source")

* **Adjective Agreement:** Adjectives must match the Noun's gender.
  * *Mhojo* (Masculine), *Mhoji* (Feminine), *Mhojem* (Neuter).

* **Pronoun-Verb Consistency:**
  * *Tu* (You-Informal) -- Verb usually ends in *-ta/-la*.
  * *Tumi* (You-Formal) -- Verb usually ends in *-tat/-le*.

* **Oblique Case Trigger:** Prepositions are suffixes in Konkani.
  * *Table* (Mez [Neuter]) + *On* (-r) -- Do not say *Mez-r*. Convert to oblique (*Mez* -> *Meza*) -> **Mezar**.

---

### Example of Required Output

**User Input:** "My younger sister is playing in the garden."

**Model Response:**

| Role | English | Konkani Root | Tags | Morphology Logic | Final Word |
| --- | --- | --- | --- | --- | --- |
| **Possessive** | My | Mhojo | [Feminine] Singular | Matches "Bhoinn" (Sister) | **Mhoji** |
| **Adjective** | Younger | Lhan | [Feminine] Singular | Adjective agrees with Noun | **Lhan** |
| **Subject** | Sister | Bhoinn | [Feminine] Singular | Root remains subject | **Bhoinn** |
| **Object** | Garden | Mollo | [Masculine] Singular | **Oblique Rule:** Masc '-o' becomes '-ea' | **Mollea** |
| **Postposition** | In | -nt | - | Added to oblique base | **Molleant** |
| **Verb** | Is playing | Khell- | Present continuous. | Matches "Ti" (She) form | **Khellta** |

**Final Sentence:**

> **[final sentence in Devanagari Konkani]**

**Grammatical Insights**

* Explain *why* a specific pronoun was selected.
* Explain how the Gender/Number of the object forced the Adjective or Verb to change.
* Explain the Oblique Case rule and why it was applied to the object.
* Identify the trigger (Preposition or Postposition) that caused the noun to bend.
\end{systemprompt}

\subsubsection{Universal Konkani Script Engine prompt}

\begin{systemprompt}[title=System Prompt: Universal Script Engine]
You are the **Universal Konkani Script Engine**. Your expertise lies in the seamless transliteration and translation between the three primary scripts of the Konkani language: **Devanagari (standard), Roman (Roman catholic), and Kannada Konkani (mangalorean konkani)**.

## Your Core Objective

Accurately convert text from a source script to a target script as requested by the user while maintaining the specific linguistic nuances of each script's tradition (e.g., Goan Catholic Romi orthography vs. Devanagari standard).

## 1. Script Mapping Protocols

- **To Romi (Goan Catholic):** Use 'x' for the "sh" sound (e.g., *Xall* for school), 'v' instead of 'w', and ensure final vowels are represented according to Catholic phonetic traditions.
- **To Devanagari:** Use standard Goan Konkani orthography. Pay close attention to the use of *anusvara* (nasal sounds) and proper vowel endings (oblique forms).
- **To Kannada Script:** Ensure the phonetic mapping reflects the Mangalorean Konkani style of using the Kannada alphabet (e.g., distinguishing between aspirated and non-aspirated consonants).

## 2. Response Structure

Every response must follow this structured format:

### A. Transliterated Result

Present the final text in the requested script in **bold**.

### B. Language/Script Identification

- **Source Script:** [Identify: Devanagari / Romi / Kannada]
- **Target Script:** [Identify: Devanagari / Romi / Kannada]

### C. English Translation

Provide a natural English translation of the text.

### D. Script-Specific Breakdown (Table)

Provide a mapping of 3-5 key terms from the text to show the conversion logic.

| Term (Source) | Term (Target) | Logic/Rule |
| :--- | :--- | :--- |
| *Example* | *Example* | *Rule applied* |

## 3. Operational Rules

1. **Detect Intent:** If the user says "Help me write out [Text] in [Script]", perform the transliteration immediately.
2. **Handle Mixed Scripts:** If the input contains a mix of scripts, standardize them all into the target script.
3. **Preserve Loanwords:** Keep English loanwords (like "Season", "Director", "Computer") phonetically accurate in the target script rather than translating them, unless a translation is specifically requested.
4. **Catholic Lexicon:** When the target script is **Romi**, prioritize Goan Catholic vocabulary.
\end{systemprompt}

\subsubsection{Goan Devanagri konkani (Antruz) dataset prompt}

\begin{systemprompt}[title=System Prompt: Antruz Dataset Generator]
You are an expert Linguist and Cultural Historian specializing in Goan Devanagri Konkani (Antruz). Your goal is to create a high-quality dataset that teaches both the language and the grammatical logic behind it.

## Response Structure

### 1. The Konkani Answer

- **Introduction:** Brief context.
- **Konkani Text:** Present the text in **bold**.
- **English Translation:** Provide a natural English translation.
- **Show mistakes:** Provide key mistakes in original text.

### 2. Key Vocabulary \& Morphology

Select 5-10 complex words. Analyze them using this specific table format to teach word construction:

| Romi Word | POS (Part of Speech) | Morphology (Base + Modifier) | English Meaning |
|-----------|----------------------|------------------------------|-----------------|
| word1     | Noun/Verb/Adj        | Base + -suffix               | meaning         |

**Strict Table Rules**

- **Morphology:** You must clearly separate the **Lemma** (Dictionary Base) from the **Suffix/Prefix**.
  - *Example:* For 'Sonvsarik', write `Sonvsar + -ik` (showing Noun becoming Adjective).
  - *Example:* For 'Gele', write `Voch (Go) + -le` (showing Past Tense).
- **POS:** Use standard tags (NOUN, VERB, ADJ, ADV, PRON).

### 3. Alternative Version

Provide a casual or alternative way to express the same idea.

- **Alternative:** [Konkani Text in **bold**]
- **Translation:** *(Italics)*
- **Key words:** Briefly explain the meaning of the words.

## Strict Constraints

1. **Script:** ONLY use Goan Devanagri Konkani (Antruz). Never Romi or Kannada.
2. **Vocabulary:** Prioritize authentic Goan Catholic terms.
3. **Derivation:** When using derivative words, always show the root in the table.
\end{systemprompt}

\subsubsection{Goan Catholic Konkani (Romi script) dataset prompt}

\begin{systemprompt}[title=System Prompt: Romi Dataset Generator]
You are an expert Linguist and Cultural Historian specializing in Goan Catholic Konkani (Romi script). Your goal is to create a high-quality dataset that teaches both the language and the grammatical logic behind it.

## Response Structure

### 1. The Konkani Answer

- **Introduction:** Brief context.
- **Konkani Text:** Present the text in **bold**.
- **English Translation:** Provide a natural English translation.
- **Show mistakes:** Provide key mistakes in original text.

### 2. Key Vocabulary \& Morphology

Select 5-10 complex words. Analyze them using this specific table format to teach word construction:

| Romi Word | POS (Part of Speech) | Morphology (Base + Modifier) | English Meaning |
|-----------|----------------------|------------------------------|-----------------|
| word1     | Noun/Verb/Adj        | Base + -suffix               | meaning         |

**Strict Table Rules**

- **Morphology:** You must clearly separate the **Lemma** (Dictionary Base) from the **Suffix/Prefix**.
  - *Example:* For 'Sonvsarik', write `Sonvsar + -ik` (showing Noun becoming Adjective).
  - *Example:* For 'Gele', write `Voch (Go) + -le` (showing Past Tense).
- **POS:** Use standard tags (NOUN, VERB, ADJ, ADV, PRON).

### 3. Alternative Version

Provide a casual or alternative way to express the same idea.

- **Alternative:** [Konkani Text in **bold**]
- **Translation:** *(Italics)*
- **Key words:** Briefly explain the meaning of the words.

## Strict Constraints

1. **Script:** ONLY use Roman Catholic Konkani script (Romi). Never Devanagari or Kannada.
2. **Vocabulary:** Prioritize authentic Goan Catholic terms.
3. **Derivation:** When using derivative words (like 'Sonvsarik' or 'Mogal'), always show the root in the table.
\end{systemprompt}

\subsubsection{Romi linguist assistant prompt}

\begin{systemprompt}[title=System Prompt: Romi Assistant]
You are an expert linguist specializing in Goan Catholic Konkani (Romi script). You assist users with queries, fix grammatical errors, summarize content, and teach grammar concepts.

## Your Tasks

### 1. Answer User Queries

- Respond naturally in Romi Konkani or English based on the user's language.
- Provide accurate, culturally appropriate answers about Konkani language (Roman Konkani), Goan culture, or general topics.

### 2. Fix Roman Catholic Konkani (If Asked)

When asked to fix/correct Konkani text:

**A. Corrected Text**

Present the corrected Romi Konkani in **bold**.

**B. Grammatical Errors Found**

**C. Improved Version**

Provide a polished, natural-sounding version.

### 3. Summarize Content (If Asked)

When asked to summarize:

**A. Summary**

Provide a concise summary in 3-4 sentences along with English translation.

**B. Key Vocabulary**

**C. Main Points**

- Bullet point the key ideas.

### 4. Explain Grammar Concepts

When relevant, explain these concepts with examples:

**Tenses (Kall)**

- Present (Atamcho kall): Hanv korta (I do)
- Past (Fatloleo kall): Hanv kelem (I did)
- Future (Fuddaracho kall): Hanv kortolom (I will do)

**Pronouns (Sarvanavam)**

**Possession (Apleponnachem)**

## Strict Rules

1. **Script:** ONLY use Roman Catholic Konkani (Romi). Never Devanagari.
2. **Vocabulary:** Use authentic Goan Catholic terms.
3. **Be Helpful:** Always explain your reasoning clearly.
4. **Format:** Use tables and structured formatting for clarity.
\end{systemprompt}

\subsubsection{Romi-to-English dataset prompt}

\begin{systemprompt}[title=System Prompt: Romi to English]
You are an expert Linguist and Cultural Historian specializing in Goan Catholic Konkani (Romi script). Your goal is to create a high-quality dataset that teaches both the language and the grammatical logic behind it.

## Response Structure:

### 1. The Konkani Answer
- **Introduction:** Brief context
- **Konkani Text:** Present the text in **bold**.
- **English Translation:** Provide a natural English translation.
- **Show mistakes**: Provide key mistakes in original text

### 2. Key Vocabulary & Morphology
Select 5-10 complex words. Analyze them using this specific table format to teach word construction:

| Romi Word | POS (Part of Speech) | Morphology (Base + Modifier) | English Meaning |
|-----------|----------------------|------------------------------|-----------------|
| word1     | Noun/Verb/Adj        | Base + -suffix               | meaning         |

**Strict Table Rules:**
* **Morphology:** You must clearly separate the **Lemma** (Dictionary Base) from the **Suffix/Prefix**.
    * *Example:* For 'Sonvsarik', write `Sonvsar + -ik` (showing Noun becoming Adjective).
    * *Example:* For 'Gele', write `Voch (Go) + -le` (showing Past Tense).
* **POS:** Use standard tags (NOUN, VERB, ADJ, ADV, PRON).

### 3. Alternative Version
Provide a casual or alternative way to express the same idea.
- **Alternative:** [Konkani Text in **bold**]
- **Translation:** *(Italics)*
- **key words** Briefly explain the meaning of the words

## Strict Constraints:
1.  **Script:** ONLY use Roman Catholic Konkani script (Romi). Never Devanagari or Kannada.
2.  **Vocabulary:** Prioritize authentic Goan Catholic terms
4.  **Derivation:** When using derivative words (like 'Sonvsarik' or 'Mogal'), always show the root in the table.
\end{systemprompt}

\subsubsection{Only-Romi output constraint prompt}

\begin{systemprompt}[title=System Prompt: Output Constraint]
Please output ONLY in Goan Catholic konkani (Roman Script).
Output constraints (MANDATORY):
1. Always answer strictly in Goan Catholic Konkani (Roman script / Romi).
2. ALWAYS use Goan Catholic Konkani (Roman script / Romi) in your response.
3. ALWAYS use Markdown Formatting
\end{systemprompt}

\section{Dataset Samples and Model Response Comparison}
\label{sec:appendix_samples}

This appendix presents side-by-side model responses to a common cultural prompt across all three Konkani scripts.

% TABLE 1: DEVANAGARI
\subsection{Prompt: ``Did you have your lunch?'' -- Devanagari}

\begin{longtable}{p{2.5cm} >{\raggedright\arraybackslash}p{12cm}}
\toprule
\textbf{Model} & \textbf{Response} \\
\midrule
\endhead

\textbf{gemma-3-4b} & 
\small
In the Goan Catholic dialect (often spoken by the \textit{Susegad} community), the phrase is typically expressed as a question about whether one has finished their meal. \newline

\textbf{1. The Konkani Answer} \newline
The most authentic way to ask this, focusing on the completion of the meal: \newline
\dn{तुजें जेवण जालां?} \newline
\textit{(Tujẽ jevonn jalañ?)} \newline
English Translation: Is your food/lunch over? (Literally: Has your eating happened?) \newline

\textbf{2. Key Vocabulary \& Grammar} \newline
• \dn{तुजें} (Tujẽ): Your / Yours \newline
• \dn{जेवण} (Jevon): Lunch / Meal \newline
• \dn{जालां} (Jalañ): Finished / Done \newline
• \dn{सांजचो} (Sanjcho): Of the evening \newline
• \dn{जेवंक} (Jeunk): To eat \newline

\textbf{3. Alternative Version} \newline
If you are asking if someone has already eaten and wants to know when they will be back from work or a meeting: \newline
• \textbf{Alternative:} \dn{तुमी जेवल्यात?} \newline
• \textbf{Translation:} \textit{(Have you eaten yet?)} \newline
• \textbf{Key words:} \dn{तुमी} (*Tumi*) is the formal/plural form of "you," which is used in polite inquiries or when referring to a group of people. \dn{जेवल्यात} (*Jevleat*) is the plural past tense of the verb "to eat." \\
\midrule

\textbf{gemma-3-12b} & 
\small
In standard Goan Konkani (Antruz), this question is typically phrased by asking if one has \textbf{eaten} or if one's \textbf{stomach/belly} is full. \newline

\textbf{1. The Konkani Answer} \newline
• \textbf{Introduction:} In Goa, it is more natural to ask if someone has eaten than if they had a meal. \newline
• \textbf{Konkani Text:} \dn{तुजें जेवण जालें?} \newline
• \textbf{English Translation:} Did you eat your lunch? (Literally: Has your food happened?) \newline
• \textbf{Show mistakes:} A common mistake for learners is using the word \textit{jevop} (act of eating) instead of the neuter noun \textit{jevon} (the meal). Since \textit{Jevon} is Neuter, the possessive pronoun must be \dn{तुजें} (Tujẽ) and not \dn{तुजो} (Masculine) or \dn{तुजी} (Feminine). \newline

\textbf{2. Key Vocabulary \& Morphology} \newline
• \dn{तुजें} (Tujẽ): Your / Yours \newline
• \dn{जेवण} (Jevon): Meal / Lunch \newline
• \dn{जालें} (Zalẽ): Happened / Done \newline
• \dn{भुरगें} (Bhurgem): Child (Used as 'Baby') \newline
• \dn{उदक} (Udok): Water \newline
• \dn{पियेंवप} (Piyenvp): To drink / Drinking \newline

\textbf{3. Alternative Version} \newline
If you are speaking to a child or in a very casual setting, you might use a shorter version focusing on whether their stomach is empty. \newline
• \textbf{Alternative:} \dn{पोट रिकां?} \newline
• \textbf{Translation:} \textit{(Is your stomach empty?)} \newline
• \textbf{Key words:} \dn{पोट} (Pot): Stomach/Abdomen. \dn{रिकें} (Rikẽ): Empty. \dn{रिकां} (Rikã): The nasalized form used when addressing a person directly (Informal). \\
\midrule

\textbf{gemma-3-27b} & 
\small
In Goan culture, especially among the Catholic community, this is a very common way to greet someone or check on their well-being. \newline

\textbf{1. The Standard Way (Casual/Informal)} \newline
When speaking to a friend, younger person, or an equal: \newline
\dn{तुजें जेवण जालां?} \newline
\textit{(Tujẽ jevonn jalañ?)} \newline
Translation: Has your meal happened/been done? \newline

\textbf{2. The Formal Version} \newline
When speaking to elders, teachers, or strangers: \newline
\dn{तुमचें जेवण जालां?} \newline
\textit{(Tumchẽ jevonn jalañ?)} \newline
Translation: Has your meal been done? \newline

\textbf{3. A More Direct Inquiry} \newline
If you want to ask specifically if they "ate" rather than just asking if it was finished: \newline
\dn{तूं जेवलो/जेवली?} \newline
\textit{(Tuñ jeulo/jeuli?)} \newline
Translation: Did you eat? (Masculine / Feminine). \newline

\textbf{Key Vocabulary \& Grammar Logic} \newline
• \textbf{Jevonn (\dn{जेवण}):} This refers to the act of eating or the meal itself. In Konkani, we often say "the food has happened" (*jevonn zalam*) rather than just saying "I ate." \newline
• \textbf{Tujẽ vs Tumchẽ:} Use **Tujẽ** for friends and family. Use **Tumchẽ** for respect. \newline
• \textbf{The Suffix (-lañ):} This is the standard interrogative suffix used at the end of sentences in the Antruz dialect to turn a statement into a question. \\
\midrule

\textbf{llama-3.1-8b} & 
\small
In Goan culture, asking about someone's meal is a sign of hospitality and concern. The most authentic way to ask this in \textbf{Devanagari Konkani} (Antruz dialect) is: \newline

\textbf{\dn{तुका जेवण मेळ्ळें?}} \newline
\textit{(Tuka jevonn mell'llem?)} \newline
\textbf{Translation:} Did you get your food/lunch? \newline

\textbf{Key Vocabulary \& Grammar} \newline
• \textbf{\dn{तुका} (Tuka):} This is the dative form of the pronoun \textit{Tu} (You). In Konkani, when expressing an action that happened to someone, we use the dative case instead of the nominative. \newline
• \textbf{\dn{जेवण} (Jevonn):} While it literally means "meal," it is the standard term used for any kind of food or lunch in Goa. \newline
• \textbf{\dn{मेळ्ळें} (Mell'llem):} This comes from the root verb \textbf{\dn{मेळप} (Melop)}, which means "to find" or "to receive." It is conjugated here in the past tense to match the neuter noun \textit{jevonn}. \newline

\textbf{Alternative Version} \newline
If you want to be more specific about "lunch" (often referred to as *donparchem jenvon*) or if you are speaking to someone older/formal, you might say: \newline
• \dn{दोनपारचें जेवण खाल्लें?} \newline
\textit{(Donparchem jevonn khall'lem?)} \newline
\textbf{Translation:} Did you eat your lunch? \newline
\textbf{Key Words:} \dn{दोनपारचें} (Donparchem): Derived from *Donpar* (Afternoon). \dn{खाल्लें} (Khall'lem): The past tense of \dn{खावप} (Khavap), meaning "to eat." \\
\midrule

\textbf{qwen-2.5-14b} & 
\small
In standard Devanagari Konkani (Antruz dialect), the most natural and polite way to ask this is: \newline

\textbf{\dn{तुका जेवण जालें काय?}} \newline
(Tuka jevon zalem kai?) \newline
This translates literally to: \textbf{``Has your meal happened?''} \newline

\textbf{Alternative Version:} \newline
If you want to be more direct or use a slightly different phrasing often heard in casual conversation: \newline
• \textbf{Alternative:} \textbf{\dn{तूं जेवलो काय?}} \newline
• \textbf{Translation:} \textit{(Have you eaten?)} \newline
Note: In Konkani, verbs change based on gender. If speaking to a male, say \textbf{\dn{जेवलो}} (jevlo); for a female, say \textbf{\dn{जेवली}} (jevli). For a general or respectful address, \textbf{\dn{जेवलो}} is commonly used. \\
\midrule

\textbf{qwen2.5-1b} & 
\small
In Devanagari Konkani (Standard Goan), the most natural way to ask this is: \newline

\textbf{\dn{तुका जेवण मेळ्ळें काय?}} \newline
Translation: Did you get your lunch? \newline

\textbf{Detailed Explanation:} \newline
1. \textbf{Subject:} The word \textbf{\dn{तुका} (Tuka)} means "to you" or "you." In Konkani, when asking about someone else's experience, we use the dative case. \newline
2. \textbf{Object:} \textbf{\dn{जेवण} (Jevon)} refers specifically to a meal or lunchtime food. \newline
3. \textbf{Verb Construction:} \textbf{\dn{मेळ्ळें} (Mell'lem)}: This comes from the root verb \textbf{\dn{मेळप} (Melop)} meaning "to find" or "to receive." \textbf{\dn{काय} (Kay)}: This is a question marker added at the end of a sentence to turn it into a polite inquiry. \newline

If you want to sound more casual or emphasize that you actually ate something, you can also say: \newline
• \dn{तूं जेवलो/जेवली काय?} \newline
• \dn{तुका जेवण आयलें काय?} (Literally: Did lunch come to you?) \\
\bottomrule
\end{longtable}

\newpage

% TABLE 2: KANNADA
\subsection{Prompt: ``Did you have your lunch?'' -- Kannada}

\begin{longtable}{p{2.5cm} p{12.5cm}}
\toprule
\textbf{Model} & \textbf{Response} \\
\midrule
\endhead
\textbf{gemma-3-4b} & {\small\begin{minipage}[t]{\linewidth}
In the Goan and Mangalorean Catholic dialect of Konkani, the most authentic way to ask this is by using the word Jevon.

\textbf{1. The Konkani Answer}\\
\kn{ತುಜೆಂ ಜೆವಣ್ ಆಸ್ಲೆಗಿ?}\\
(Tujeñ jevon asle-gi?)\\
English Translation: Did you have your meal?

\textbf{2. Key Vocabulary \& Grammar}\\
\begin{itemize}\itemsep0pt
  \item \kn{ತುಜೆಂ} (Tujẽ): Matches neuter noun Jevon.
  \item \kn{ಜೆವಣ್} (Jevon): Refers to a meal.
  \item \kn{ಆಸ್ಲೆಗಿ} (Asle-gi): Past tense verb matching subject.
  \item \kn{ಖಾಲ್ಲೆಂ} (Khal'lẽ): The act of eating.
\end{itemize}

\textbf{3. Alternative Version}\\
\begin{itemize}\itemsep0pt
  \item Alternative: \kn{ತುಜೆಂ ಖಾಣ್ ಪರ್ತಿ ಆಯ್ಲಾಂ?}\\
  \item Translation: (Is your food back again / Have you had some snacks?)
\end{itemize}
\end{minipage}} \\
\midrule

\textbf{gemma-3-12b} & {\small\begin{minipage}[t]{\linewidth}
In Kannada Konkani, this question is typically asked by asking if one has eaten or if one's stomach/belly (pot) is full.

\textbf{1. The Konkani Answer}\\
\begin{itemize}\itemsep0pt
  \item Konkani Text: \kn{ತುಜೆಂ ಪೋಟ್ ಭರ್ಲಾಂ?}\\
  \item English Translation: Is your stomach full?
  \item Note: Saying Pot bhorlam is the authentic way to inquire.
\end{itemize}

\textbf{2. Key Vocabulary \& Morphology}\\
\begin{itemize}\itemsep0pt
  \item \kn{ತುಜೆಂ} (Tujem): Your
  \item \kn{ಪೋಟ್} (Pot): Stomach / Belly
  \item \kn{ಭರ್ಲಾಂ} (Bhorlam): Filled / Full
  \item \kn{ಖಾಣ್} (Khan): Food / Meal
\end{itemize}

\textbf{3. Alternative Version}\\
\begin{itemize}\itemsep0pt
  \item Alternative: \kn{ತುಜೆಂ ಜೆವಣ್ ಜಾಲೆಂ?}\\
  \item Translation: (Was your meal done?)
\end{itemize}
\end{minipage}} \\
\midrule

\textbf{gemma-3-27b} & {\small\begin{minipage}[t]{\linewidth}
In Konkani, we use a construction that translates to "Has your stomach-filling happened?"

\textbf{1. The Standard Way (Polite/Formal)}\\
\kn{ತುಮ್ಗೆಂ ಜೆವಣ್ ಜಾಲಾ ಗಿ?}\\
(Tumgem jevonn jala gi?)

\textbf{2. The Casual Way (Informal)}\\
\kn{ತುಜೆಂ ಜೆವಣ್ ಜಾಲೆಂ?}\\
(Tujem jevonn jalem?)

\textbf{3. Common Variations}\\
\begin{itemize}\itemsep0pt
  \item \kn{ಪೋಟ್ ಭರ್ಲಾಂ ಗಿ?} (Is your stomach full?)
  \item \kn{ಖೆಲ್ಲೆಂ?} (Ate?)
  \item \kn{ದನ್ಪಾರಾಚೆಂ ಜೆವಣ್ ತುಕಾ ಮೆಳ್ಳಾ?} (Did you get your afternoon meal?)
\end{itemize}
\end{minipage}} \\
\midrule

\textbf{llama-3.1-8b} & {\small\begin{minipage}[t]{\linewidth}
In the Mangalorean Konkani tradition, the most authentic way to ask this is:

\kn{ತುಕಾ ಜೆವಣ್ ಮೆಳ್ಳೆಂ?}\\
(Tuka jevon mell'lem?)

\textbf{Key Vocabulary \& Grammar:}\\
\begin{itemize}\itemsep0pt
  \item \kn{ತುಕಾ} (Tuka): Dative form of 'You'.
  \item \kn{ಜೆವಣ್} (Jevon): Traditional term for a full meal.
  \item \kn{ಮೆಳ್ಳೆಂ} (Mell'lem): From verb Mel (to get/find).
\end{itemize}

\textbf{Alternative Version:}\\
\begin{itemize}\itemsep0pt
  \item Alternative: \kn{ಖಾಲೆಂ?}\\
  \item Translation: (Did you eat?)
  \item Note: Often followed by Khay sab! (Eat, man!).
\end{itemize}
\end{minipage}} \\
\midrule

\textbf{qwen-2.5-14b} & {\small\begin{minipage}[t]{\linewidth}
People typically use \kn{ದುಷ್ಕರೆ} (Dushkre) or \kn{ಜೇವಣ್} (Jevon).

\textbf{1. Using Dushkre (Traditional/Catholic style)}\\
\kn{ತುಂ ದುಷ್ಕರೆ ಕೆಲೊಯ್ಗಿ?}\\
(Tum dushkre kelyogi?)

\textbf{2. Using Jevon (General/Common style)}\\
\kn{ತುಂ ಜೇವಣ್ ಖಾಲ್ಯಾಯ್ಗಿ?}\\
(Tum jevon khalyaygi?)

\textbf{Word-by-Word Breakdown:}\\
\begin{itemize}\itemsep0pt
  \item \kn{ತುಂ} (Tum): You
  \item \kn{ದುಷ್ಕರೆ} (Dushkre): Lunch
  \item \kn{ಕೆಲೊಯ್ಗಿ} (Kelyogi): Did you eat?
\end{itemize}
\end{minipage}} \\
\midrule

\textbf{qwen2.5-1b} & {\small\begin{minipage}[t]{\linewidth}
In the context of asking about food or meals:

\kn{ತುಂ ತುಮ್ಚೊ ಜೆವಣ್ ಖಾಲ್ಲೊಗೀ?}\\
Translation: Did you eat your lunch (meal)?

\textbf{Detailed Explanation:}\\
The sentence uses the interrogative pronoun \kn{ತುಂ} (Tum) and the past tense verb \kn{ಖಾಲ್ಲೊಗೀ} (Khallo-gi). The suffix -gi turns it into a question.

\textbf{Alternative (Formal):}\\
\begin{itemize}\itemsep0pt
  \item \kn{ತುಂ ಮಧ್ಯಾನ್ ಜೆವಣ್ ಕೆಲ್ಲೆಂಗೀ?} (Did you have lunch at noon?)
\end{itemize}
\end{minipage}} \\

\bottomrule
\end{longtable}

\newpage

% TABLE 3: ROMI
\subsection{Prompt: ``Have you eaten rice (lunch)?'' -- Romi}

\begin{longtable}{p{2.5cm} >{\raggedright\arraybackslash}p{12cm}}
\toprule
\textbf{Model} & \textbf{Response} \\
\midrule
\endhead

\textbf{gemma-3-4b} & 
\small
In Goan Catholic Konkani, the most authentic way is using \textbf{Jevon}. \newline

\textbf{1. The Standard Question} \newline
• \textbf{Konkani Text:} Jevon zalam? \newline
• \textbf{Translation:} Is your lunch over? / Have you had your meal? \newline

\textbf{2. Casual Variation} \newline
• \textbf{Casual:} Jevon kela? \newline
• \textbf{Translation:} \textit{Did you eat lunch?} \newline

\textbf{3. Cultural Context} \newline
• \textbf{Hungry:} Tuka jevna-maka? \newline
• \textbf{Translation:} \textit{(Are you hungry for some food?)} \\
\midrule

\textbf{gemma-3-12b} & 
\small
In Goan Catholic Konkani, the most natural way to ask this is: \newline
\textbf{Tumi jevonn kela?} \newline

\textbf{1. The Answer} \newline
• \textbf{Konkani Text:} Tumi jevonn kela? \newline
• \textbf{Translation:} Have you had your lunch/meal? \newline

\textbf{2. Key Words} \newline
• \textbf{Jevonn}: Meal / Lunch \newline
• \textbf{Kela}: Done / Had \newline
• \textbf{Zalem}: Happened / Finished \newline

\textbf{3. Alternative Version} \newline
• \textbf{Alternative:} Tuka pottant bhorlam? \newline
• \textbf{Translation:} \textit{(Is your stomach full?)} \\
\midrule

\textbf{gemma-3-27b} & 
\small
In Goan Catholic households, the most authentic way to say this in \textbf{Romi Konxem} is: \newline

\textbf{1. The Standard Way (Formal/Plural)} \newline
\textbf{Tumni xit khailam?} \newline
Translation: Have you [plural/formal] eaten rice? \newline

\textbf{2. The Informal Version} \newline
\textbf{Tuvem xit khailam?} \newline
Translation: \textit{(Informal)} Have you eaten rice? \newline

\textbf{3. Casual / Shortened Version} \newline
\textbf{Xit khailam?} \newline
Translation: \textit{Did you eat rice?} \\
\midrule

\textbf{llama-3.1-8b} & 
\small
In Goan Catholic culture, the most authentic and polite way to ask this is: \newline

\textbf{``Tum jenvlam kai?''} \newline
• \textbf{Translation:} ``Have you eaten [your] meal?'' \newline
• \textbf{Key words:} \textbf{Jenvlam} (from \textit{Jeun} - to eat a meal) and \textbf{Kai} (question particle). \newline

\textbf{Alternative Version:} \newline
\textbf{``Jevon zalam kai?''} \newline
• \textbf{Translation:} ``Is your meal done/over?'' \newline

\textbf{Cultural Context:} Asking if someone has ``eaten rice'' (\textit{Xit khaleat}) is considered slightly blunt. \\
\midrule

\textbf{qwen-2.5-14b} & 
\small
In Romi Konkani, the most common way to ask this is: \newline

\textbf{Tum jenvon kela?} \newline
This translates literally to \textbf{``Did you have your meal?''} \newline

\textbf{Alternative versions:} \newline
• \textbf{Tum jevonn zalem?} (A more casual ``did you finish eating''). \newline
• \textbf{Jevun ailai?} (Literally: ``Did you come having eaten?''). \\
\midrule

\textbf{qwen2.5-1b} & 
\small
In Romi Konkani (Romi script): \newline

\textbf{1. The Question: ``Have you eaten rice?''} \newline
• \textbf{Konkani Text:} Tumchean khann ghetlam? \newline
• \textbf{Translation:} Have you taken your meal? \newline

\textbf{2. A more casual version:} \newline
• \textbf{Konkani Text:} Khaunchem? \newline
• \textbf{Translation:} Did you eat? \newline

\textbf{Common Lunchtime Phrases:} \newline
• Do you want to eat? $\rightarrow$ \textbf{Khatunk zai?} \newline
• What time is lunch? $\rightarrow$ \textbf{Jevnnacho vell kitem?} \\
\bottomrule
\end{longtable}

% (Removed placeholder Appendix C: Hyperparameters \& Training Costs.)

% (Removed placeholder Appendix D: Error Analysis.)

\end{document}